%% file: blind_score.tex
\documentclass[british]{article}
\usepackage[T1]{fontenc}
\usepackage[latin9]{inputenc}
\PassOptionsToPackage{natbib=true}{biblatex}
\usepackage{xcolor}
\usepackage{refstyle}
\usepackage{float}
\usepackage{amsmath}
\usepackage{amsthm}
\usepackage{amssymb}
\usepackage{graphicx}

\makeatletter

\AtBeginDocument{\providecommand\figref[1]{\ref{fig:#1}}}
\AtBeginDocument{\providecommand\exaref[1]{\ref{exa:#1}}}
\AtBeginDocument{\providecommand\lemref[1]{\ref{lem:#1}}}
\AtBeginDocument{\providecommand\propref[1]{\ref{prop:#1}}}
\AtBeginDocument{\providecommand\subsecref[1]{\ref{subsec:#1}}}
\RS@ifundefined{subsecref}
  {\newref{subsec}{name = \RSsectxt}}
  {}
\RS@ifundefined{thmref}
  {\def\RSthmtxt{theorem~}\newref{thm}{name = \RSthmtxt}}
  {}
\RS@ifundefined{lemref}
  {\def\RSlemtxt{lemma~}\newref{lem}{name = \RSlemtxt}}
  {}

\theoremstyle{definition}
    \ifx\thechapter\undefined
      \newtheorem{example}{\protect\examplename}
    \else
      \newtheorem{example}{\protect\examplename}[chapter]
    \fi
\theoremstyle{plain}
    \ifx\thechapter\undefined
      \newtheorem{lem}{\protect\lemmaname}
    \else
      \newtheorem{lem}{\protect\lemmaname}[chapter]
    \fi
\theoremstyle{plain}
    \ifx\thechapter\undefined
      \newtheorem{prop}{\protect\propositionname}
    \else
      \newtheorem{prop}{\protect\propositionname}[chapter]
    \fi
\theoremstyle{remark}
    \ifx\thechapter\undefined
      \newtheorem{rem}{\protect\remarkname}
    \else
      \newtheorem{rem}{\protect\remarkname}[chapter]
    \fi

\usepackage[nonatbib,final]{neurips_robustbayes21}
\author{Li K. Wenliang~~~~~~~~~~Heishiro Kanagawa \\   
Gatsby Computational Neuroscience Unit\\
University College London\\   
London, UK \\   
\texttt{kevinwli@outlook.com~~~heishiro.kanagawa@gmail.com}}
\newref{exa}{%
    name      = Example~,  
    names     = Example~,
    Name      = Example~,
    Names     = Example~,
    rngtxt    = Example~,
    lsttwotxt = Example~,
    lsttxt    = Example~ 
}
\newref{fig}{%
    name      = \RSFigtxt,  
    names     = \RSFigstxt,
    Name      = \RSFigtxt,
    Names     = \RSFigstxt,
    rngtxt    = \RSrngtxt,
    lsttwotxt = \RSlsttwotxt,
    lsttxt    = \RSlsttxt
}

\newref{lem}{%
    name      = Lemma~,  
    names     = Lemmas~,
}

\newref{prop}{%
    name      = Proposition~,  
    names     = Propositions~,
}
\newref{claim}{%
    name      = Claim~,  
    names     = Claim~,
}

\newref{sec}{%
    name      = Section~,  
    names     = Section~,
}

\newref{subsec}{%
    name      = Section~,  
    names     = Section~,
}

\usepackage{hyperref}

\usepackage{xcolor}
\hypersetup{
    colorlinks,
    linkcolor={red!50!black},
    citecolor={blue!50!black},
    urlcolor={blue!80!black}
}

\makeatother

\usepackage{babel}
\usepackage[bibstyle=numeric,sorting=ynt,maxbibnames=99]{biblatex}
\providecommand{\examplename}{Example}
\providecommand{\lemmaname}{Lemma}
\providecommand{\propositionname}{Proposition}
\providecommand{\remarkname}{Remark}

\begin{document}
\include{math_commands}

\title{Blindness of score-based methods to isolated components and mixing
proportions}
\maketitle
\begin{abstract}
Statistical tasks such as density estimation and approximate Bayesian
inference often involve densities with unknown normalising constants.
Score-based methods, including score matching, are popular techniques
as they are free of normalising constants. Although these methods
enjoy theoretical guarantees, a little-known fact is that they exhibit
practical failure modes when the unnormalised distribution of interest
has isolated components --- they cannot discover isolated components
or identify the correct mixing proportions between components. We
demonstrate these findings using simple distributions and present
heuristic attempts to address these issues. We hope to bring the attention
of theoreticians and practitioners to these issues when developing
new algorithms and applications.
\end{abstract}

\section{Introduction and background}

This paper presents a pervasive practical issue of score-based methods.
The (Hyvärinen) score function of a differentiable probability density
$p(x)$ is defined by $s_{p}(x):=\nabla_{x}p(x)/p(x).$ Score function
does not depend on the normaliser and, therefore, has a broad range
of applications in machine learning and Bayesian statistics. Chief
among these are the following: (a) training unnormalised density models
with score matching (SM) \citep{HyvaerinenHyvaerinen2005Estimation},
(b) measuring the quality of approximate samplers using Stein discrepancies
(SDs) \citep{GorhamMackey2015Measuring,ChwialkowskiGretton2016Kernel,LiuJordan2016Kernelized,GorhamMackey2017Measuring,GorhamMackey2019Measuring},
and (c) approximate posterior sampling via Stein variational gradient
descent (SVGD) \citep{LiuWang2016Stein}. We show that these theoretically
well-motivated score-based algorithms can fail in practice when the
unnormalised distribution has \emph{isolated components}. In what
follows, we exemplify the common failure modes with the following
simple setup:
\begin{example}[Gaussian mixtures]
 Define the following density functions on $\sR$: \label{exa:gmm}
\begin{gather*}
p(x)=\pi_{1}p_{1}(x)+(1-\pi_{1})p_{2}(x),\,\,q(x)=p_{1}(x),\,\,p_{1}(x)=\gN(x;-\mu,\sigma^{2}),\,p_{2}(x)=\gN(x;\mu,\sigma^{2}).
\end{gather*}
where $\mu,\sigma>0$, and $\pi_{1}\in(0,1)$ are mixing proportions.
In addition, we define $p'(x):=\pi_{1}'p_{1}(x)+(1-\pi_{1}')p_{2}(x)$
as the same mixture as $p(x)$ but with a different mixing proportion
$\pi_{1}'\ne\pi_{1}$. Instances of these densities are shown in \figref{fisher}.
When $\mu/\sigma^{2}$ is large, the components of $p$ are \emph{isolated}.
\begin{figure}[t]
\begin{centering}
\includegraphics[width=1\textwidth]{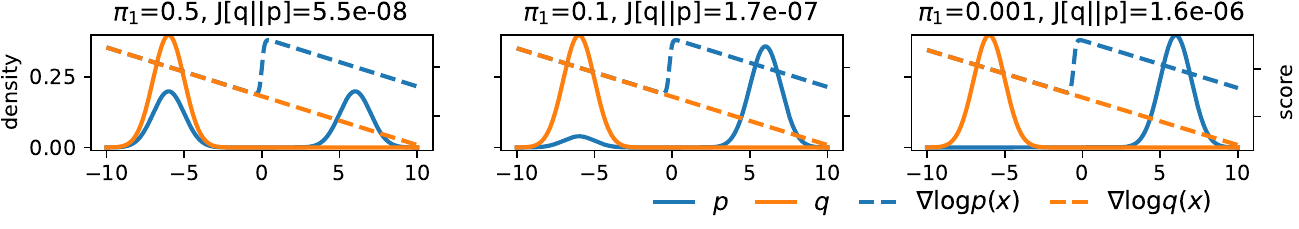}
\par\end{centering}
\begin{centering}
\includegraphics[width=1\textwidth]{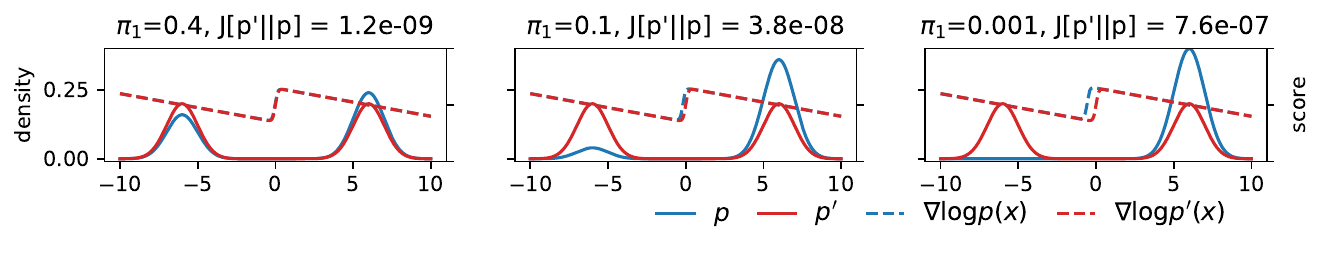}\vspace{-1.5em}
\par\end{centering}
\centering{}\caption{Distributions in \exaref{gmm} and their Fisher divergence (FD, $J$)
for various choices of $\pi_{1}$ for the mixture \textcolor{blue}{$p$}
(panel titles). Top, $J\!\left[{\color{orange}q}\|{\color{blue}p}\right]$
is blind to the presence of an isolated component in the mixture \textcolor{blue}{$p$}
regardless of $\pi_{1}$. Bottom, $J\!\left[{\color{red}p'}\|{\color{blue}p}\right]$
is blind to different mixing proportions. \label{fig:fisher}\vspace{-0.5em}}
\end{figure}

The aforementioned failure modes stem from the following Lemma concerning
the distributions in \exaref{gmm}, the proof of which can be found
in Appendix \ref{subsec:lemma-1-proof}.
\end{example}
\begin{lem}[Weak dependence of $s_{p}$ on $\pi_{1}$]
 For the densities $p$ and $q$ defined in \exaref{gmm}, $s_{p}(x)$
gets arbitrarily close to $s_{p_{1}}(x)=s_{q}(x)$ regardless of $\pi_{1}$
for $x\ne0$ when $\mu/\sigma^{2}$ gets large.\label{lem:weak_dependence}
\end{lem}
This property is illustrated in \figref{fisher} (top) --- the score
$s_{p}(x)$ does not change visibly with $\pi_{1}.$ In the following
sections, we discuss the consequences of \lemref{weak_dependence}
for the three applications introduced above. To our knowledge, there
has not been a synthesised exposition of the common failure modes,
except for references \citep{ArbelGretton2018gradient,ZhuoZhang2018Message,WenliangGretton2019Learning,SongErmon2019Generative,GorhamMackey2019Measuring,ZhangChen2020Stochastic,DAngeloFortuin2021Annealed}
which on specific algorithms or other issues. We also propose heuristic
remedies to initiate an effort to rectify these issues.

\section{Fisher divergence and Stein discrepancy}

Consider training an unnormalised density model $\tilde{p}(\vx)$
with score $s_{p}$ on data drawn from $q(x).$ \citet{HyvaerinenHyvaerinen2005Estimation}
proposed SM to train $\tilde{p}$ by minimising the Fisher divergence
(FD) 
\[
J[q||p]=\frac{1}{2}\int q(x)\left\Vert s_{q}(x)-s_{p}(x)\right\Vert _{2}^{2}\ud x.
\]
The FD is zero if and only if $p=q$. The densities $p$ and $q,$
however, can still be ``very different'' when their FD is close
to but not exactly zero, as we show below (see also Appendix \ref{subsec:Proof-of-weakdep}).
\begin{prop}[FD is blind to isolated components]
 For $q$ and $p$ in \exaref{gmm}, the FD $J[q||p]\to0$ regardless
of the mixing proportion $\pi_{1}$ when \textup{$\mu/\sigma^{2}$}
gets large.\label{prop:spurious_mode}\vspace{-0.5em}
\end{prop}
\emph{Proof sketch.~} The FD $J[q||p]$ is an expectation under $q$
which has almost all of its mass in $x<0$ as $\mu/\sigma^{2}$ gets
large. \lemref{weak_dependence} implies that $s_{p}(x)\to s_{p_{1}}(x)=s_{q}(x)$
for $x<0$. Thus, $J[q||p]\to0.$

Another issue arises when two mixtures $p'$ and $p$ comprise the
same set of components weighted by different mixing proportions. In
this case, their FD is almost zero despite the large difference in
term of probability mass. 
\begin{prop}[FD is blind to $\pi$]
 For $p$(x) and $p'(x)$ defined by distinct choices of the mixing
proportion in \exaref{gmm}, the FD $J(p'\|p)\to0$ as $\mu/\sigma^{2}$
gets large regardless of the mixing proportion. \label{prop:sm_mixing_prop}\vspace{-1.6em}
\end{prop}
\emph{Proof sketch.~} By \lemref{weak_dependence}, the scores $s_{p}$
and $s_{p'}$ converge to the same limit for $x\ne0.$ They differ
substantially only for $x$ close to 0 where $p'$ puts vanishing
mass, so $J(p'\|p)\to0$.

In density estimation where $p$ is the model, \propref{spurious_mode}
implies that the model can have a mixture component far away from
the data $q.$ According to \propref{sm_mixing_prop}, when isolated
components in a data distribution $p'$ are well-fit individually
by the model $p$, one can obtain another model with small FD by varying
the mixing proportions arbitrarily. See \figref{fisher} (bottom row)
for visualisations. We discuss in \subsecref{score-matching-success}
how previous successes of SM avoided these issues.

Next, we discuss the score-based Stein discrepancies (SDs) \citep{GorhamMackey2015Measuring,ChwialkowskiGretton2016Kernel,LiuJordan2016Kernelized,GorhamMackey2019Measuring}
which can measure how well samples from $q$ agree with model $\tilde{p}$.
A (Langevin) SD between $q$ and $p$ is defined as
\begin{align}
\SDf\left[q\|p\right] & =\sup_{f\in{\cal F}}\left|\E{x\sim q}{s_{p}(x)^{\top}f(x)+\nabla_{x}^{\top}f(x)}\right|,\label{Stein}
\end{align}
where $\nabla_{x}^{\top}$ is the divergence operator, and $\gF$
is a class of differentiable vector-valued functions with appropriate
boundary conditions (see the foregoing references for precise definitions).
Since the SD is upper bounded by the FD (see Appendix \ref{subsec:Proof-of_blind_stein}),
we have the following:
\begin{prop}[Blindnesses of SD]
 For $f\in\gF$ such that $\int\Vert f(x)\Vert_{2}^{2}q(x)dx\leq1$
and $q(x)f(x)\to0$ as $\Vert x\Vert_{2}\to\infty$, $\mathrm{SD}_{\gF}\!\left[q\|p\right]$
and $\mathrm{SD}_{\gF}\!\left[p'\|p\right]$ suffer from the issues
of $J\left[q\|p\right]$ and $J\left[p'\|p\right]$ in \propref[s]{spurious_mode}
and \ref{prop:sm_mixing_prop}, respectively.\label{prop:blind_stein}
\end{prop}
In the case of the kernel SD (KSD), where $\gF$ is the unit ball
of a reproducing kernel Hilbert space (RKHS) \citep{ChwialkowskiGretton2016Kernel,LiuJordan2016Kernelized},
we show in \figref{ksd} that the best $f\in\gF$ witnessing the difference
between $p$ and $q$ is almost zero around $x=0$. Therefore, diagnostics
based on KSD can be misleading. We further discuss this issue in \subsecref{sd_bounds_ipm}
by relating to KSD bounds on integral probability metrics. 
\begin{figure}
\begin{centering}
\includegraphics[width=1\textwidth]{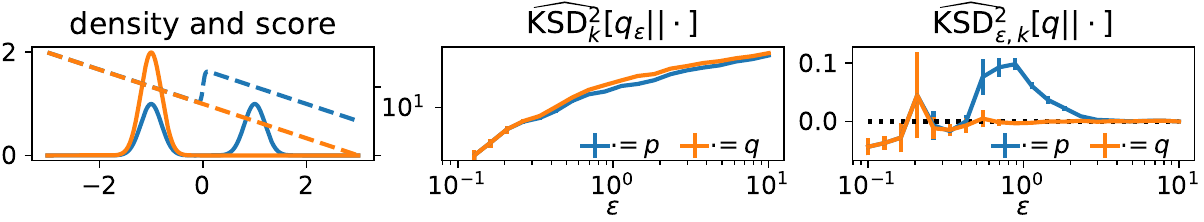}
\par\end{centering}
\begin{centering}
\includegraphics[width=1\textwidth]{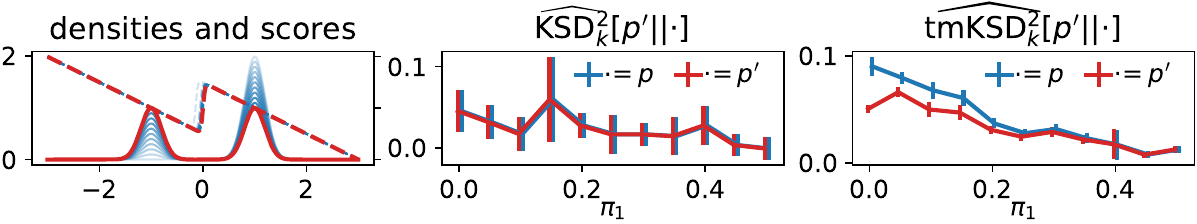}\vspace{-1em}
\par\end{centering}
\caption{Top: densities (left) in \exaref{gmm} and their squared KSD (middle)
and tmKSD (right) estimated with 500 samples. Bottom: the squared
KSD (middle) and tmKSD (right) between ${\color{red}p'}$ and a few
${\color{blue}p}$ with various $\pi_{1}$ estimated with 2000 samples.
Errorbar shows 1 s.e. estimated by Jackknife.\label{fig:tmKSD}}
\end{figure}

\paragraph{Heuristic remedy for KSD: matching a noisy $q$ and a tempered $p$}

To better detect isolated components,consider adding noise to $q$
and changing the temperature of $p$:
\[
q_{\epsilon}(x):=\int q(x')\gN(x|x',\epsilon^{2})\ud x'\text{ where }\epsilon>0,\quad\tilde{p}^{\beta}(x)\propto\bigl(p(x)\bigr)^{\beta}\textrm{ where }\beta\in(0,1].
\]
The KSD between $q_{\epsilon}$ and $\tilde{p}^{\beta}$ is given
in closed-form \citep{ChwialkowskiGretton2016Kernel,LiuJordan2016Kernelized}:
\vspace*{-0.1em}
\begin{align}
\textrm{KSD}_{k}^{2}[q_{\epsilon}\|\tilde{p}^{\beta}] & =\beta^{2}\E{x,x'\sim q_{\epsilon}}{s_{p}(x)^{\top}s_{p}(x')k(x,x')}+\beta\E{x,x'\sim q_{\epsilon}}{s_{p}(x)^{\top}\nabla_{x'}k(x,x')}\nonumber \\
 & \quad\,+\beta\E{x,x'\sim q_{\epsilon}}{s_{p}(x')^{\top}\nabla_{x}k(x,x')}+\E{x,x'\sim q_{\epsilon}}{\mathrm{tr}[\nabla_{x}\nabla_{x'}k(x,x')]},\label{eq:ksd_k_quadratic}
\end{align}
with $\mathrm{tr}$ the matrix trace. Adding noise to $q$ will likely
increase this KSD, but we can adjust the temperature in $p$ to compensate
for this effect, since both transformations ``broaden'' the original
densities. By noting that (\ref{eq:ksd_k_quadratic}) is convex and
quadratic in $\beta$ for positive-definite $k$, we take its unique
infimum over $\beta$ to counter the noise-induced mismatch, giving
\begin{equation}
\KSDke[q\|p]:=\textrm{KSD}_{k}^{2}[q_{\epsilon}\|\tilde{p}^{\beta^{*}\!(\epsilon)}]\text{,\ensuremath{\quad}where\quad}\beta^{*}\!(\epsilon)=\argmin_{\beta\in(0,1]}\textrm{KSD}_{k}^{2}[q_{\epsilon}\|\tilde{p}^{\beta}].\label{eq:tm-KSD}
\end{equation}
For the densities defined in \exaref{gmm}, we visualise $\KSDke\left[q\|p\right]$
as a function of $\epsilon$ and compare it with $\textrm{KSD}_{k}^{2}[q_{\epsilon}\|p]$
without temperature matching in \figref{tmKSD} (top). For some values
of $\epsilon$ $\KSDke[q\|p]$ is significantly greater than the baseline
$\KSDke[q\|q]$ (right), but this is not the case for $\textrm{KSD}_{k}^{2}[q_{\epsilon}\|\cdot]$
(middle), suggesting the importance of matching the temperature with
noise. Second, $\KSDke$ reaches a maximum at some $\epsilon$. We
define this maximum as the temperature-matched KSD
\[
\textrm{tmKSD}_{k}^{2}\left[q\|p\right]:=\max_{\epsilon}\KSDke\left[q\|p\right].
\]
Note that this is not a valid discrepancy (see Appendix \ref{sec:further-matching}),
but it is more sensitive to isolated components than KSD. We show
these SDs between a fixed $p'$ and a few $p$ with various $\pi_{1}$
in \figref{tmKSD} (bottom). Although $\textrm{tmKSD}_{k}^{2}\left[p'\|p\right]$
still cannot reliably identify the correct mixing proportion 0.5,
its smaller estimation errors and stronger dependence on $\pi_{1}$
compared to KSD are encouraging. We show very similar results for
non-Gaussian distributions in Appendix \ref{sec:further-matching}.

\section{Stein variational gradient descent}

\begin{figure}[t]
\begin{centering}
\includegraphics[width=1\textwidth]{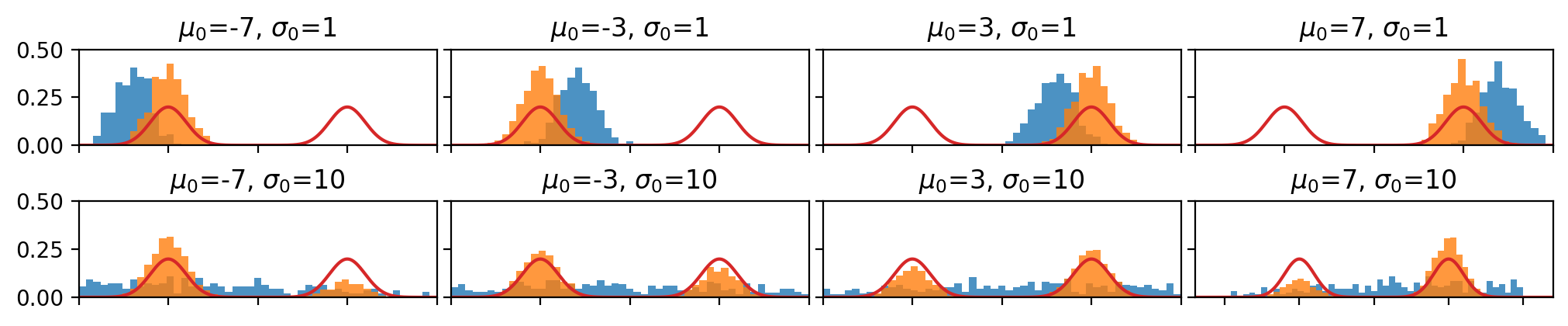}\vspace{-1em}
\par\end{centering}
\caption{SVGD fails in approximating the target density $p$ (red) in \exaref{gmm}.
The initial $\nu_{0}$ is Gaussian $\protect\gN(\mu_{0},\sigma_{0}^{2})$
(blue) with small (top) or large (bottom) variances. Orange is the
final $\nu_{t}$. \label{fig:gmm_svgd}}
\end{figure}
\begin{figure}[t]
\begin{centering}
\includegraphics[width=1\textwidth]{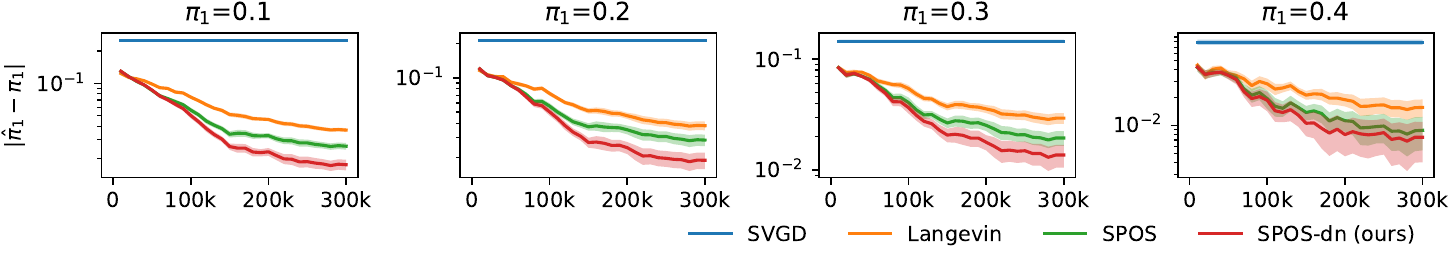}\vspace{-1em}
\par\end{centering}
\caption{The difference between the fraction of particles $<0$ $(\hat{\pi}_{1})$
and the true mixing proportion ($\pi_{1}$) when running SVGD, LS,
SPOS and the proposed SPOS-dn to approximate $p$ in \exaref{gmm}.
Lines are mean $\pm$ se estimated from 20 independent runs. See Appendix
\ref{sec:noisy-svgd} for more results. \label{fig:spos_d}}
\end{figure}
SVGD \citep{LiuWang2016Stein} approximates an unnormalised distribution
$p$ by iteratively updating an empirical distribution $\nu_{t}$
formed by particles. The main idea is to find a direction $\phi$
such that the particle update $x\leftarrow x+\epsilon\phi(x)$ lowers
the Kullback-Leibler divergence $\KL[\nu_{t}\|p]$. For $\phi$ defined
by a function in the RKHS associated with a kernel $k(\cdot,\cdot)$,
the optimal $\phi^{*}$ for a particle $x'$ is given by 
\begin{equation}
\phi^{*}(x')=\E{x\sim\nu_{t}}{s_{p}(x)k(x,x')+\nabla_{x}k(x,x')}.\label{eq:svgd}
\end{equation}
The particles can be initialised as samples drawn from a simple distribution
$\nu_{0}$, such as a Gaussian with mean $m_{0}$ and variance $\sigma_{0}^{2}$. 
\begin{prop}[Blindness of SVGD]
 For the mixture $p$ in \exaref{gmm}, if $\nu_{t}=q$ or $\nu_{t}=p'$,
one has that $\left\Vert \phi^{*}(x')\right\Vert _{2}\to0$ as $\mu/\sigma^{2}$
gets large. \label{prop:blind_svgd}

\vspace{-0.5em}
\end{prop}
\emph{Proof sketch. }By the reproducing formula \citep[Sec. 4.2]{Steinwart2008},
$\left\Vert \phi^{*}(x')\right\Vert _{2}\leq\sqrt{k(x',x')}\SDf\left[\nu_{t}\|p\right]$,
with ${\cal F}$ the unit ball of the RKHS. The upper bound goes to
zero as $\mu/\sigma^{2}$ gets large by \propref{blind_stein}.

This means that the particles get stuck at $q$ or $p'$, missing
a component in $p$ completely or giving a wrong mixing proportions.
We verify this empirically with results shown in \figref{gmm_svgd}.
The final $\nu_{t}$ is highly sensitive to the initial $\nu_{0}$;
contrary to intuitions, an overdispersed $\nu_{0}$ alone does not
help. \vspace{-0.5em}

\paragraph{Heuristic remedy: combine SVGD and Langevin sampling}

Langevin sampling (LS) targets the same stationary distribution as
SVGD while being more exploratory. A heuristic strategy is thus to
update the particles according to a combination of LS and SVGD:
\begin{equation}
x'\leftarrow x'+\epsilon_{1}\E{\nu_{t}}{s_{p}(x)k(x,x')+\nabla_{x}k(x,x')}+\epsilon_{2}s_{p}(x')+\sqrt{2\epsilon_{2}}\omega,\quad\epsilon_{1},\epsilon_{2}\ge0,\ \epsilon_{1}\epsilon_{2}\ne0,
\end{equation}
where $\omega$ is the standard normal. We let $\epsilon_{2}$ decrease
gradually while keeping $\epsilon_{1}$ fixed. This is similar to
SPOS \citep{ZhangChen2020Stochastic} which reduces both $\epsilon_{1}$
and $\epsilon_{2}$, so we refer to our heuristic by SPOS-dn (decreasing
the noisy LS step). We ran LS, SVGD, SPOS and SPOS-dn to sample $p$
in \exaref{gmm} and estimated the final mixing proportions. \figref{spos_d}
shows that SPOS and SPOS-dn are better than LS and SVGD, and SPOS-dn
converges the fastest. Nonetheless, finding the correct $\pi_{1}$
is still challenging for all algorithms tested. Details and additional
results showing robustness to $\nu_{0}$ are in Appendix \ref{sec:noisy-svgd}. 

Another approach proposed by \citet{DAngeloFortuin2021Annealed} is
to run SVGD while annealing $p$. However, unlike adding noise, annealing
does not preserve the masses of isolated components (see Appendix
\ref{subsec:annealing}). Thus, this method still produces wrong mixing
proportions \citep[Fig. 4]{DAngeloFortuin2021Annealed}.

\section{Discussion\label{subsec:modes_disscussion}}

We have demonstrated that three popular score-based methods fail to
detect the isolated components or to identify the correct mixing proportions.
The heuristic remedies presented here encourage more principled solutions.
We stress that the practical failure modes presented here do not diminish,
in any way, the theoretical advances of score-based methods; these
methods have empowered practitioners to tackle a variety of statistical
problems involving intractable distributions with unknown normalisers.
Further, the issues discussed here may or may not impact certain downstream
applications. For example, the model estimated by SM may still be
suitable for local gradient-based methods, such as denoising. In contrast,
unconditioned gradient-based sampling over the whole support may suffer
from these issues. When using SVGD for Bayesian neural networks, ignoring
the trivial posterior components arising from the symmetry of the
weights does not affect the predictive distribution. We discuss other
score-based methods that do not suffer from these issues in Appendix
\ref{subsec:entropy_gradient}.

\paragraph{Acknowledgement}

We thank Maneesh Sahani and Arthur Gretton for useful discussions.
This research is funded by the Gatsby Charitable Foundation. 

\printbibliography

\newpage{}

\appendix

\section{Proof}

For notational simplicity we define $\pi_{2}:=1-\pi_{1}$.

\subsection{Proof of \lemref{weak_dependence}\label{subsec:lemma-1-proof}}

The score function is 
\begin{align*}
\nabla_{x}\log p(x) & =\frac{\pi_{1}\nabla_{x}p_{1}(x)+\pi_{2}\nabla_{x}p_{2}(x)}{\pi_{1}p_{1}(x)+\pi_{2}p_{2}(x)}\\
 & =-\frac{\pi_{1}\exp\left[-\frac{\left(x+\mu\right)^{2}}{2\sigma^{2}}\right]\frac{\left(x+\mu\right)}{\sigma^{2}}+\pi_{2}\exp\left[-\frac{\left(x-\mu\right)^{2}}{2\sigma^{2}}\right]\frac{\left(x-\mu\right)}{\sigma^{2}}}{\pi_{1}\exp\left[-\frac{\left(x+\mu\right)^{2}}{2\sigma^{2}}\right]+\pi_{2}\exp\left[-\frac{\left(x-\mu\right)^{2}}{2\sigma^{2}}\right]}\\
 & =-\frac{\pi_{1}\frac{\left(x+\mu\right)}{\sigma^{2}}+\pi_{2}\exp\left[\frac{\left(x+\mu\right)^{2}}{2\sigma^{2}}-\frac{\left(x-\mu\right)^{2}}{2\sigma^{2}}\right]\frac{\left(x-\mu\right)}{\sigma^{2}}}{\pi_{1}+\pi_{2}\exp\left[\frac{\left(x+\mu\right)^{2}}{2\sigma^{2}}-\frac{\left(x-\mu\right)^{2}}{2\sigma^{2}}\right]}\\
 & =-\frac{\pi_{1}\frac{\left(x+\mu\right)}{\sigma^{2}}+\pi_{2}\exp\left[2{\color{blue}\frac{\mu}{\sigma^{2}}}x\right]\frac{\left(x-\mu\right)}{\sigma^{2}}}{\pi_{1}+\pi_{2}\exp\left[2{\color{blue}{\color{blue}\frac{\mu}{\sigma^{2}}}}x\right]}\\
 & =-\left(\frac{x}{\sigma^{2}}+\frac{\pi_{1}-\pi_{2}\exp\left[2{\color{blue}\frac{\mu}{\sigma^{2}}}x\right]}{\pi_{1}+\pi_{2}\exp\left[2{\color{blue}{\color{blue}\frac{\mu}{\sigma^{2}}}}x\right]}\frac{\mu}{\sigma^{2}}\right).
\end{align*}
In the limit of $\mu,\sigma^{2}$ such that ${\color{blue}\mu/\sigma^{2}}\to\infty$,
we can see that 
\begin{align*}
\left|\nabla_{x}\log p(x)-\left(-\frac{x+\mu}{\sigma^{2}}\right)\right|\to0\  & \text{for}\ x<0;\\
\left|\nabla_{x}\log p(x)-\left(-\frac{x-\mu}{\sigma^{2}}\right)\right|\to0\  & \text{for}\ x>0.
\end{align*}
The two limits can be identified with $\nabla_{x}\log p_{1}(x)=-(x+\mu)/\sigma^{2}$
for $x<0$ and $\nabla_{x}\log p_{2}(x)=-(x-\mu)/\sigma^{2}$ for
$x>0$ . 
\begin{rem}
A similar result holds for an arbitrary number of components without
the Gaussian assumption. Consider a mixture of $K$ components with
conditional likelihoods $p(x|z=k)$ for $k\in\{1,\dots,K\}$ that
may differ across components. We say that the components are isolated
if $p(z|x)$ is concentrated on a single component for all data points.
In this case, we can write $s_{p}(x)$ as 
\[
s_{p}(x)=\frac{\nabla_{x}\int p(x|z)p(z)\ud z}{p(x)}=\frac{\int\nabla_{x}\log p(x|z)p(x|z)p(z)\ud z}{p(x)}=\sum_{k=1}^{K}p(z=k|x)\nabla_{x}\log p(x|z=k).
\]
For a given $x$, if the posterior is concentrated on $z=k$, then
it is clear that $s_{p}$ is approximately equal to $\nabla_{x}\log p(x|z=k)$,
the score of the $k$th component.
\end{rem}

\subsection{Proof of \propref{spurious_mode}\label{subsec:Proof-of-weakdep}}

We split the integral in the definition of the Fisher divergence into
the positive and negative parts. For the positive part, by the definition
of $q,$ we have

\begin{align*}
\int_{0}^{\infty}q(x)\bigl(s_{q}(x)-s_{p}(x)\bigr)^{2}\ud x & =\int_{0}^{\infty}q(x)\bigl(s_{p}(x)-s_{p_{1}}(x)\bigr)^{2}\ud x.
\end{align*}
From the proof of \lemref{weak_dependence}, one can check that 
\begin{align*}
\int q(x)\bigl(s_{p}(x)-s_{p_{1}}(x)\bigr)^{2}\ud x & =4\int q(x)\left(\frac{\pi_{2}\mu/\sigma^{2}}{\pi_{1}\exp\left[2x\mu/\sigma^{2}\right]+\pi_{2}}\right)^{2}\ud x.
\end{align*}
Since $\mu/\sigma^{2}>0$, for each point $x\in(0,\infty)$, the integrand
converges to $0$ as $\mu/\sigma^{2}\to\infty$ and is bounded as
\[
\left(\frac{\pi_{2}\mu/\sigma^{2}}{\pi_{1}\exp\left[2x\mu/\sigma^{2}\right]+\pi_{2}}\right)^{2}\leq\left(\frac{1}{2x}\frac{W\bigl(\pi_{2}/\pi_{1}\cdot e^{-1}\bigr)+1}{(\pi_{1}/\pi_{2})\exp\left\{ W\bigl(\pi_{2}/\pi_{1}\cdot e^{-1}\bigr)+2x\right\} +1}\right)^{2},
\]
with $W$ the Lambert W function. The upper bound is integrable with
respect to the distribution $q.$ Thus, by the dominated convergence
theorem, we have 
\[
\int_{0}^{\infty}q(x)\bigl(s_{p}(x)-s_{p_{1}}(x)\bigr)^{2}\ud x\to0\ \text{as}\ \frac{\mu}{\sigma^{2}}\to\infty.
\]
The same conclusion can be similarly shown for the negative part,
and therefore we have $J[q||p]\to0$ regardless of the mixing proportion
$\pi_{1}$ when $\mu/\sigma^{2}\to\infty$.

\subsection{Proof of \propref{blind_stein}\label{subsec:Proof-of_blind_stein}}

Under the stated class $\gF$, observe that 
\begin{align*}
S_{\gF}\left[q\|p\right] & =\sup_{f\in{\cal F}}\left|\E q{s_{p}(x)^{\top}f(x)+\nabla_{x}^{\top}f(x)}\right|\\
 & =\sup_{f\in{\cal F}}\left|\E q{\{s_{p}(x)-s_{q}(x)\}^{\top}f(x)}\right|\\
 & \leq\sup_{\{f:\int f(x)^{2}q(x)dx\leq1\}}\left|\E q{\{s_{p}(x)-s_{q}(x)\}^{\top}f(x)}\right|\\
 & \leq\sqrt{\int\Vert s_{p}(x)-s_{q}(x)\Vert_{2}^{2}q(x)dx}\sup_{\{f:\int\lVert f(x)\rVert_{2}^{2}q(x)dx\leq1\}}\sqrt{\int\Vert f(x)\Vert_{2}^{2}q(x)dx}\\
 & \leq\sqrt{2J[q\Vert p]}.
\end{align*}
The second line is by integration by parts, the third line is from
the integral assumption on ${\cal F}$, and the fourth line follows
from the Cauchy-Schwartz inequality. As $J[q\Vert p]$ tends to zero
as $\mu/\sigma^{2}\to\infty,$ so does the lower bound $S_{\gF}\left[q\|p\right].$

\begin{figure}[H]
\begin{centering}
\includegraphics[width=0.8\textwidth]{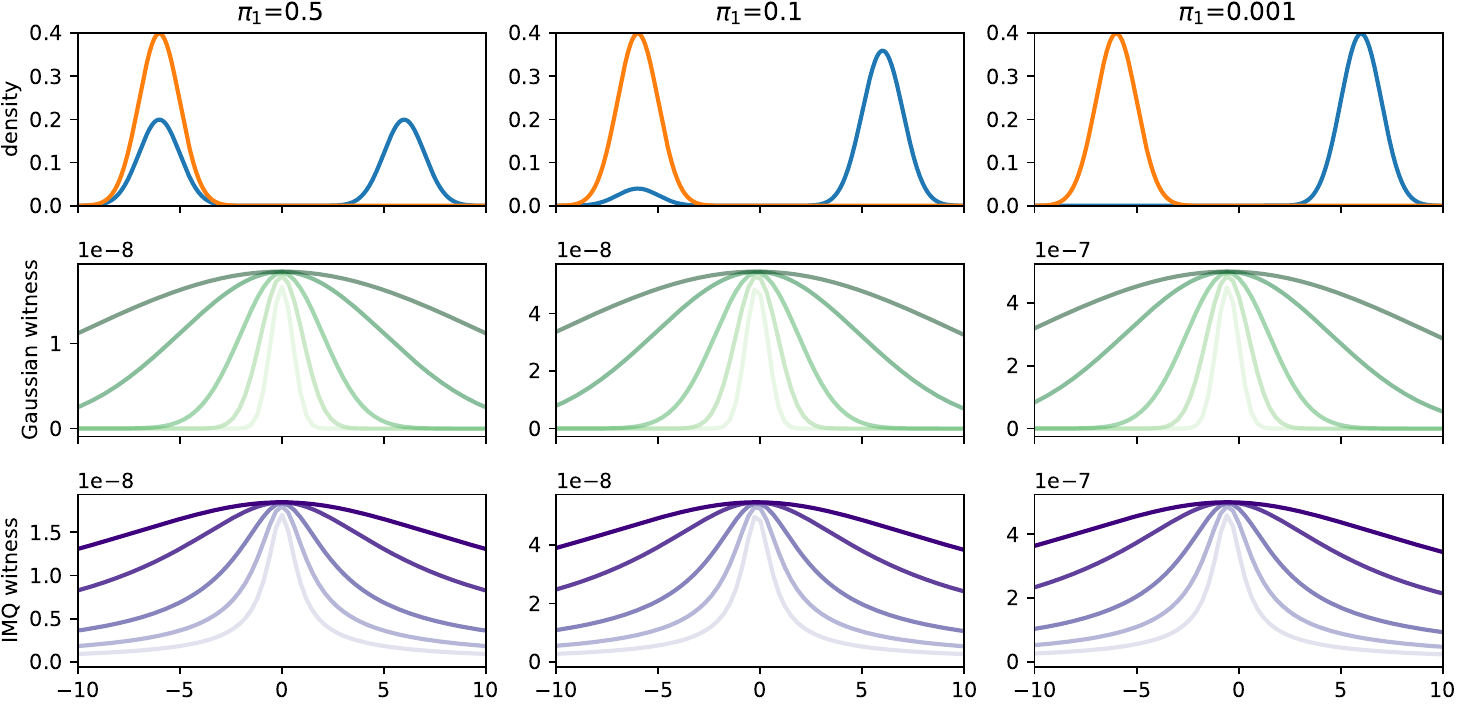}
\par\end{centering}
\caption{Top row, densities $p$ (blue) and $q$ (orange). Middle and bottom
rows show witness functions for $\textrm{KSD}\left[q\|p\right]$ given
by Gaussian and IMQ kernels, respectively. Darker colour indicates
wider bandwidths of the kernels.\label{fig:ksd}}
\end{figure}

\begin{figure}[H]
\begin{centering}
\includegraphics[width=0.8\textwidth]{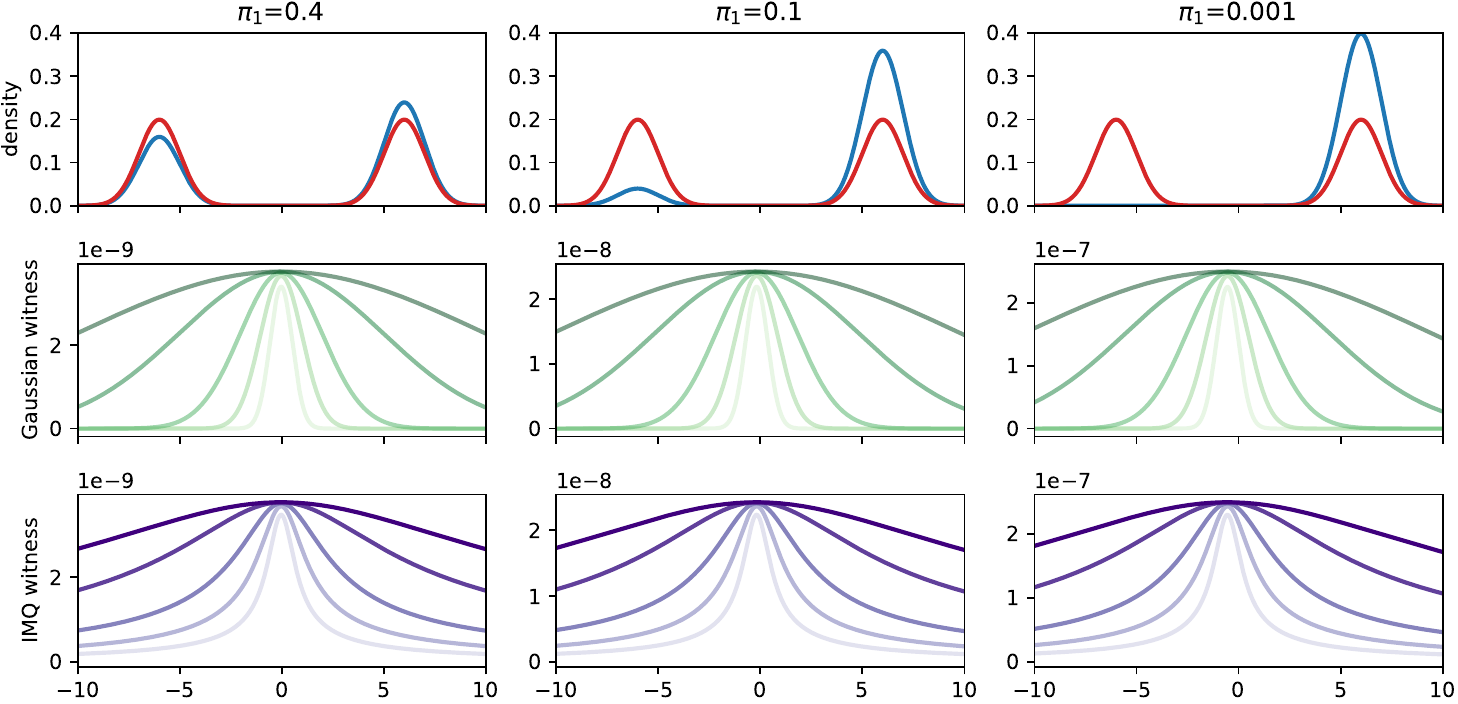}
\par\end{centering}
\caption{Same as \figref{ksd} but for $\textrm{KSD}\left[p'\|p\right]$ where
$p'$ (red) has a fixed mixing proportion $0.5$ and $p$ (blue) has
mixing proportions indicated on at the top.\label{fig:ksd_mix}}
\end{figure}

To provide more intuition, we computed the optimal $f$ (witness function)
for densities in \exaref{gmm} when $\gF$ is given by the RKHS associated
with a kernel $k$ (KSD)
\[
\E q{s_{p}(x)^{\top}k(x,\cdot)+\nabla_{x}^{\top}k(x,\cdot)}=\E q{\{s_{p}(x)-s_{q}(x)\}^{\top}k(x,\cdot)}.
\]
We repeated this with both the Gaussian and IMQ kernels \citep{GorhamMackey2017Measuring}
with various bandwidths (0.5, 1.0, 2.0, 5.0 and 10.0). The results
are shown in \figref{ksd} in \figref{ksd} for $q$ and $p$ and
\figref{ksd_mix} for $p$ and $p'$ with in \figref{ksd_mix}. For
all kernels and bandwidths, the witness functions are almost zero. 

\subsection{Proof of \propref{blind_svgd}}

For $x\in R^{d}$ and a reproducing kernel $k$ with associated RKHS
$\gH$, define 
\[
\xi_{i}(x,\cdot):=\E{x\sim\nu_{t}}{s_{p,i}(x)k(x,\cdot)+\nabla_{x_{i}}k(x,\cdot)},\quad i\in\{1,\dots,d\}
\]
By the reproducing property and Cauchy-Schwarz, we have
\begin{align*}
\left\Vert \phi^{*}(x')\right\Vert _{2}^{2} & =\sum_{i=1}^{d}\xi_{i}^{2}(x,x')=\sum_{i=1}^{d}\left\langle k(x',\cdot),\xi_{i}(x,\cdot)\right\rangle _{\gH}^{2}\\
 & \le\left\Vert k(x',\cdot)\right\Vert _{\gH}^{2}\sum_{i=1}^{d}\left\Vert \xi_{i}(x,\cdot)\right\Vert _{\gH}^{2}=k(x',x')\mathrm{KSD}^{2}(\nu_{t}\|p),
\end{align*}
where we followed the definition of $\gH^{d}$ and KSD in \citep{ChwialkowskiGretton2016Kernel}.

\section{Temperature-matched Kernel Stein Discrepancy \label{sec:further-matching}}

\subsection{Relative magnitude of tmKSD}

In the example given in \figref{tmKSD}, we obtained a zero tmKSD
between a Gaussian $q$ and itself. This is because adding zero-mean
Gaussian noise to and changing the temperature of a Gaussian $q$
both yield another Gaussian, so it is always possible to find a value
of $\beta$ such that $q_{\beta}=q_{\epsilon}$, giving a zero tmKSD
as desired. If two Gaussian distributions $q$ and $p$ differ only
in their variance, then KSD can easily detect the difference, while
tmKSD cannot. Thus, tmKSD is better used to find specifically for
isolated components after the usual KSD test.

More generally, when distributions are not restricted to Gaussians,
then adding noise to and changing the temperature of the same distribution
may result in nonzero KSDs. Thus, tmKSD is not a proper metric, and
the absolute magnitude may not be indicative of goodness-of-fit. However,
we see empirically that $\KSDke[q\|q]$ is lower than $\KSDke[q\|p]$
when $q\ne p$, which suggests that tmKSD may be used as a relative
measure.

\subsection{Experiments on other mixture distributions}

\begin{figure}
\begin{centering}
\includegraphics[width=1\textwidth]{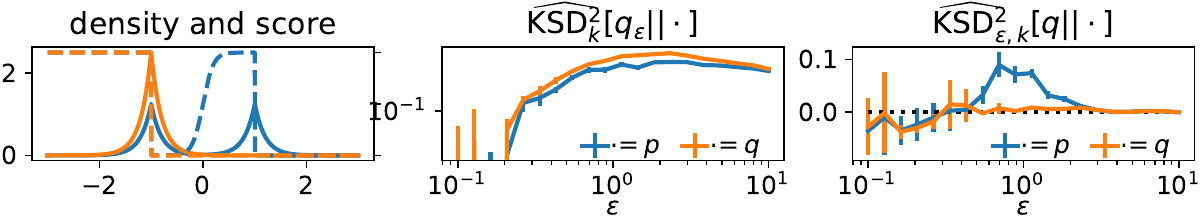}
\par\end{centering}
\begin{centering}
\includegraphics[width=1\textwidth]{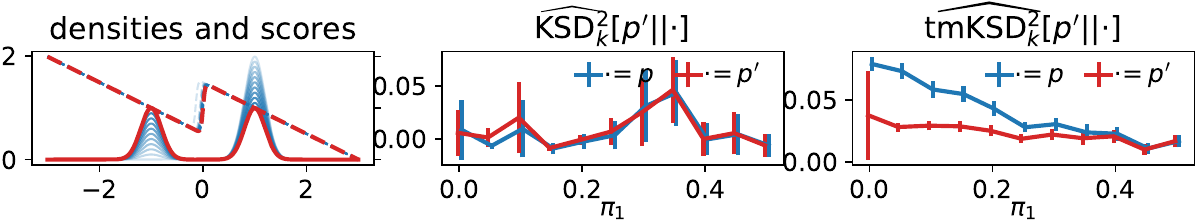}
\par\end{centering}
\caption{Same as \figref{tmKSD} but for Laplace distribution and its mixtures.
\label{fig:tmKSD_MoL}}
\end{figure}

\begin{figure}
\begin{centering}
\includegraphics[width=1\textwidth]{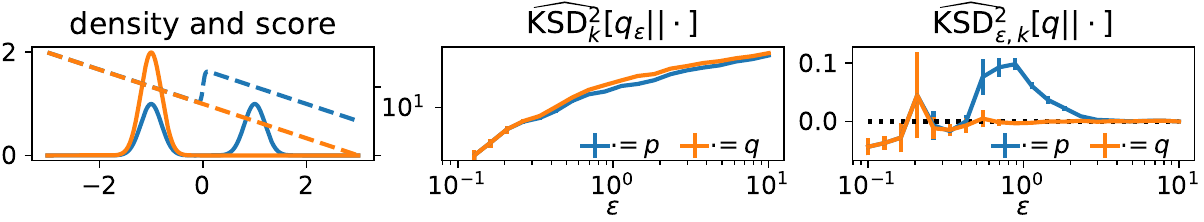}
\par\end{centering}
\begin{centering}
\includegraphics[width=1\textwidth]{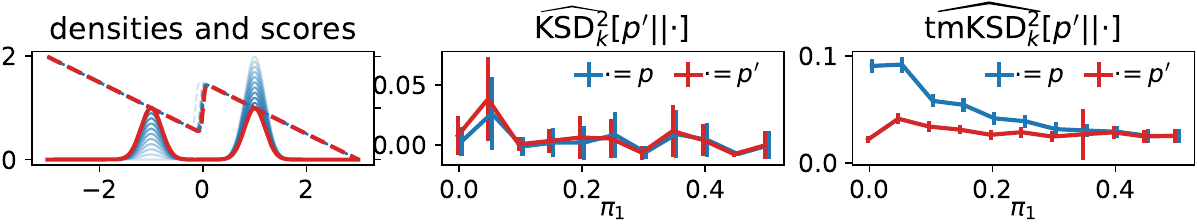}
\par\end{centering}
\caption{Same as \figref{tmKSD} but for student-T distribution and its mixtures.
\label{fig:tmKSD_MoS}}
\end{figure}

To further validate the proposed tmKSD, we ran additional experiments
on Laplace and Student-t distributions, which have heavier tails and
more dispersed samples. The same experimental procedures as \figref{tmKSD}
are used here. The results for Laplace distributions are shown in
\figref{tmKSD_MoL}, and those for Student-t (d.o.f 5.0) in \figref{tmKSD_MoS}.
They are largely consistent with the results on the Gaussian distributions.
Note that, for these distributions, the squared tmKSD between $q$
and $p$ is never smaller than that between $q$ and itself.

\section{Combining SVGD with LS \label{sec:noisy-svgd}}

Empirically, we found that SVGD gave good solutions even with $\epsilon_{1}$
fixed at 1.0, so $\epsilon_{1}$ does not need to be decreased. Intuitively,
LS contributes by mixing the initial particles to explore for isolated
components, and SVGD ``fine-tunes'' the final particles thanks to
the coupling between different particles. 

We report the procedure and hyperparameters used for these experiments.
For SVGD, we used $\epsilon=1.0$. For LS, we reduced the step size
linearly from 1.0 to 0.0 at steps of 0.01. For the original SPOS,
we reduced both $\epsilon_{1}$ and $\epsilon_{2}$ using the same
linear schedule. For our SPOS-dn, we applied this linear schedule
only to $\epsilon_{2}$ while keeping $\epsilon_{1}$ fixed at 1.0.
All kernels used are squared-exponential with unit bandwidth. For
each mixing proportion, we repeated each algorithm 20 times with different
random seeds. To evaluate the effective mixing proportion of the final
particles, we calculated the fraction of samples below 0.0 as $\hat{\pi}_{1}$
and report the error 
\begin{equation}
\Delta\pi_{1}:=\left|\hat{\pi}_{1}-\pi_{1}\right|\label{eq:delta_pi}
\end{equation}
where $\pi_{1}$ is the true mixing proportion in $p$. 

\begin{figure}
\begin{centering}
\includegraphics[width=1\textwidth]{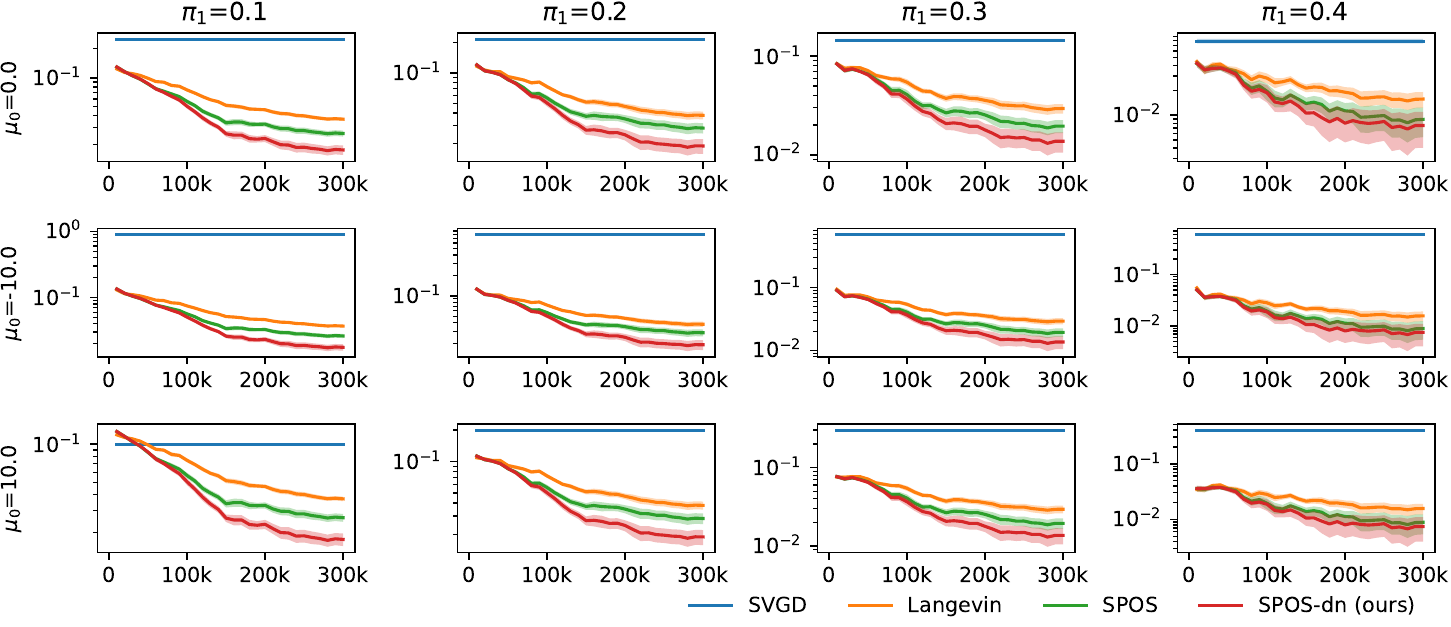}
\par\end{centering}
\caption{The trajectory of $\Delta\pi_{1}$ in (\ref{eq:delta_pi}) for SVGD,
LS, SPOS and the proposed SPOS-dn to approximate the density $p$
in \exaref{gmm}. Three rows are results from initial particles distributed
as Gaussians with the indicated means and unit variance. Lines are
mean $\pm$ se estimated from 20 independent runs.\label{fig:spos_d_all}}
\end{figure}

The results in \figref{spos_d} were obtained when the initial distribution
is $\nu_{0}=\gN(0,1)$. We ran more simulations with $\nu_{0}$ sampled
from $\gN(-10.0,1)$ or $\gN(10.0,1)$ and report all results in \figref{spos_d_all}.
In all settings, the proposed SPOS-dn gave the best estimated mixing
proportions. 

\section{Further discussions\label{sec:Further-discussions}}

\subsection{Previous successes with score matching \label{subsec:score-matching-success}}

Previous successes on energy-based models required additional constraints
preprocessing. To build a full probabilistic density model that supports
all downstream statistical applications (e.g. density estimation,
empirical Bayes, parameter interpretation, etc.), the issue of \propref{spurious_mode}
can be partially alleviated by controlling the tail behaviour of $p$
\citep{ArbelGretton2018Kernel,WenliangGretton2019Learning}, although
the notion of tails in a mixture distribution with isolated components
may be harder to define. The issue of \propref{sm_mixing_prop} can
be partially addressed by fitting to each component after clustering
the data \citep{WenliangGretton2019Learning}. \citet{SongErmon2019Generative}
initiated a novel approach to training a sequence of score functions
for sample generation, which gave very impressive results. However,
this approach is not yet a generic learning algorithm for any given
energy-based model for full downstream applications. More explicitly,
the method of training a sequence of score functions is at odds with
training a single energy-based model with a fixed architecture.

\subsection{Isolated modes and Stein discrepancy bounds on integral probability
metrics \label{subsec:sd_bounds_ipm}}

Diffusion-based Stein discrepancies are known to upper-bound integral
probability metrics (IPMs) such as the $L^{1}$-Wasserstein distance
\citep{GorhamMackey2015Measuring,GorhamMackey2019Measuring} or the
Dudley metric \citep{GorhamMackey2017Measuring}. The key assumption
in those results is that the diffusion has a fast Wasserstein decay
rate as detailed in Section 2.2 of \citep{GorhamMackey2019Measuring};
dissipativity conditions are sufficient for this requirement \citep[Section 3]{GorhamMackey2019Measuring}.
A Gaussian mixture with a fixed shared variance satisfies the distant
dissipativity condition. The $L^{1}$-Wasserstein rate of a diffusion
targeting the distribution, however, can be slow, as shown in Proposition
3.4 of \citep{Erdogdu2018}; the rate has a factor exponential in
the maximum distance between modes. Therefore, constants (known as
Stein factors) appearing in the upper-bounds in the aforementioned
papers can be large, and thus a small Stein discrepancy value might
not imply the closeness in the IPM. This observation reflects the
blindness of Stein discrepancies \propref{blind_stein} -- two Gaussian
mixtures with largely different mixture weights should have different
means and hence a large value of the $L^{1}$-Wasserstein distance. 

\subsection{Annealing does not preserve probability mass\label{subsec:annealing}}

\citet{DAngeloFortuin2021Annealed} recently introduced this idea
to SVGD and produced better samples. However, the mixing proportions
are still incorrectly estimated. This is because annealing cannot
preserve the mixing proportions between different temperatures. 

\begin{figure}
\begin{centering}
\includegraphics[width=1\textwidth]{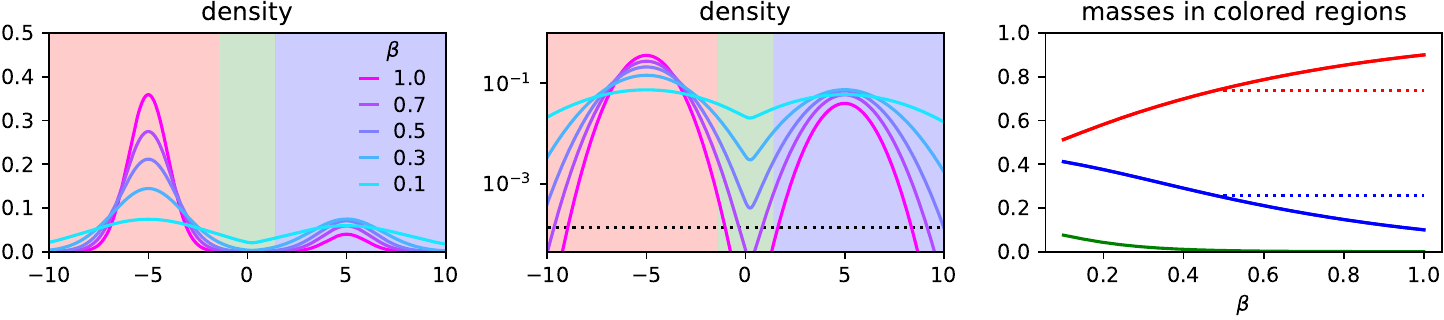}
\par\end{centering}
\caption{Left: densities of a Gaussian mixture at different temperatures $\beta$.
Middle: log normalised densities of the mixtures. The dashed line
delineates a threshold such that densities below this threshold is
considered low, and very few samples exist. For illustrative purpose,
this threshold is taken as the Gaussian density evaluated at 4 standard
deviations from the mean. During annealing, the density at $x=0$
falls below the threshold as $\beta$ exceeds 0.5. Right, the mass
in the middle green region becomes negligible when $\beta>0.5$ even
when the masses in the red and blue regions are changing substantially.
This means that during annealing there are hardly any samples that
would appear in or transition through the green region. The relative
sample proportion as $\beta$ increase beyond 0.5 is almost the same
as when $\beta=0.5$, following the dotted lines, while the true proportions
continue to change, following the solid lines. \label{fig:gaussian_temp}}
\end{figure}

Consider the mixture of two Gaussian with density shown in \figref{gaussian_temp}
(left). The probability masses round the two components change substantially
as temperature varies, and they stay isolated. We manually pick a
density threshold (middle, dashed black) below which we consider as
low-density. When $\beta<0.5$, the density at $x=0.0$ is above the
threshold for the support shown, and there are no isolated components;
SVGD will correctly sample the distribution. When $\beta>0.5$, the
components become isolated as the density at $x=0$ falls below the
threshold in the green region (right), but the masses of the two components
in the red and blue regions are still converging slowly to masses
in the original mixture when $\beta=1$. Thus, when the particles
are well-mixed at $\beta=0.5$, the proportions of particles allocated
to $x<0$ and $x>0$ do not agree with the correct mixing proportions.
The wrong mixing proportion are carried over into lower temperature
up to $\beta=1$ (right, dotted lines). Note that this happens regardless
of the annealing schedule. 

\subsection{Entropy gradient estimation does not suffer from the blindness\label{subsec:entropy_gradient}}

The score function appears in estimating the gradient of entropy of
implicit distributions, e.g.\ \citep{LiTurner2018Gradient,ShiZhu2018Spectral}.
Consider an implicit distribution $p_{\phi}(x)$ defined by the mapping
$f_{\phi}:z\mapsto x$, $z\sim\zeta$ where $\zeta$ is some simple
distribution and $f_{\phi}$ is a flexible function parametrised by
$\phi$. The gradient of the entropy satisfies
\[
\nabla_{\phi}\sH[p_{\phi}(x)]=\E{\zeta\left(z\right)}{\nabla_{x}\log p_{\phi}(x)\nabla_{\phi}f_{\phi}(z)}.
\]
There are two reasons why this application does not suffer from the
blindness discussed here. First, samples from $p(x)$ can be easily
drawn from the implicit distribution, unlike when $p$ is an energy-based
model. Second, the expectation above is an expectation of a $p_{\phi}(x)$-dependent
function under $p_{\phi}(x)$ itself (through $\zeta$), which cannot
be blind to itself. This is unlike SM or SD where the expectation
involves two different density functions $p(x)$ and $q(x)$.
\end{document}

%% file: math_commands.tex
\global\long\def\vectorfont#1{\boldsymbol{#1}}%

\global\long\def\randomvectorfont#1{#1}%

\global\long\def\tensorfont#1{\boldsymbol{\mathsf{#1}}}%

\global\long\def\setfont#1{\mathbb{#1}}%

\global\long\def\matrixfont#1{\mathbf{#1}}%

\global\long\def\graphfont#1{\mathcal{#1}}%

\global\long\def\mi{\text{\text{-}}}%

\global\long\def\vx{\vectorfont x}%
\global\long\def\x{\vx}%

\global\long\def\vy{\vectorfont y}%
\global\long\def\y{\vy}%

\global\long\def\vz{\vectorfont z}%
\global\long\def\z{\vz}%
\global\long\def\zm{\vz_{m}}%

\global\long\def\vw{\vectorfont w}%
\global\long\def\w{\vw}%

\global\long\def\vphi{\vectorfont{\phi}}%
\global\long\def\valpha{\vectorfont{\alpha}}%
\global\long\def\vpsi{\vectorfont{\psi}}%
\global\long\def\vf{\vectorfont f}%
\global\long\def\vtheta{\vectorfont{\theta}}%
\global\long\def\vgamma{\vectorfont{\gamma}}%
\global\long\def\vepsilon{\vectorfont{\epsilon}}%
\global\long\def\vb{\vectorfont b}%
\global\long\def\vpi{\vectorfont{\pi}}%
\global\long\def\softplus{\textrm{softplus}}%
\global\long\def\vm{\vectorfont m}%
\global\long\def\vmu{\vectorfont{\mu}}%
\global\long\def\vr{\vectorfont r}%
\global\long\def\va{\vectorfont a}%
\global\long\def\vp{\vectorfont p}%
\global\long\def\vh{\vectorfont h}%
\global\long\def\vs{\vectorfont s}%
\global\long\def\vrb{\bar{\vectorfont r}}%
\global\long\def\vu{\vectorfont u}%
\global\long\def\vlambda{\vectorfont{\lambda}}%
\global\long\def\vomega{\vectorfont{\omega}}%
\global\long\def\viota{\vectorfont{\iota}}%
\global\long\def\vc{\vectorfont c}%
\global\long\def\vn{\vectorfont n}%
\global\long\def\vzeta{\vectorfont{\zeta}}%
\global\long\def\vsigma{\vectorfont{\sigma}}%
\global\long\def\vrho{\vectorfont{\rho}}%
\global\long\def\vv{\vectorfont v}%
\global\long\def\vg{\vectorfont g}%
\global\long\def\veta{\vectorfont{\eta}}%
\global\long\def\vbeta{\vectorfont{\beta}}%
\global\long\def\vk{\vectorfont k}%
\global\long\def\vl{\vectorfont l}%
\global\long\def\vo{\vectorfont o}%
\global\long\def\ve{\vectorfont e}%

\global\long\def\vzt{\vectorfont z_{t}}%
\global\long\def\vxt{\vectorfont x_{t}}%
\global\long\def\vzot{\vectorfont z_{1:t}}%
\global\long\def\vxot{\vectorfont x_{1:t}}%
\global\long\def\vztm{\vectorfont z_{t\text{-}1}}%
\global\long\def\vxtm{\vectorfont x_{1\text{-}t}}%
\global\long\def\vztp{\vectorfont z_{t+1}}%
\global\long\def\vxtp{\vectorfont x_{1\text{+}t}}%
\global\long\def\vxotm{\vectorfont x_{1:1\text{-}t}}%
\global\long\def\vzotm{\vz_{1:1\text{-}t}}%
\global\long\def\vrt{\vectorfont r_{t}}%
\global\long\def\vrtm{\vectorfont r_{t\mi1}}%
\global\long\def\vpsit{\vpsi_{t}}%
\global\long\def\vxoT{\vectorfont x_{1:T}}%
\global\long\def\vrtT{\vectorfont r_{t|1:T}}%
\global\long\def\vrtpT{\vectorfont r_{t+1|1:T}}%
\global\long\def\vrtmtau{\vectorfont r_{t\mi\tau}}%
\global\long\def\vrtmtau{\vectorfont r_{t\mi\tau}}%
\global\long\def\vztmtau{\vz_{t\mi\tau}}%
\global\long\def\ntheta{\vectorfont{\eta}_{\vtheta}}%
\global\long\def\lnorm{\Phi}%
\global\long\def\ud{\mathrm{d}}%
\global\long\def\xn{\vx_{n}}%
\global\long\def\hatJ{\hat{J}}%

\global\long\def\ptheta{p_{\vtheta}}%
\global\long\def\nat{\vectorfont{\eta}}%
\global\long\def\norm{\Phi}%
\global\long\def\suff{\vectorfont s}%
\global\long\def\ntheta{\nabla_{\vtheta}}%
\global\long\def\pdata{p_{0}}%
\global\long\def\ptilde{\tilde{p}}%
\global\long\def\lptheta{\log\ptheta}%
\global\long\def\nlptheta{\nabla_{\vtheta}\lptheta}%
\global\long\def\ltheta{\Phi_{\vtheta}}%
\global\long\def\pthetat{p_{\vtheta_{t}}}%
\global\long\def\qphi{q_{\vphi}}%
\global\long\def\nlpjoint{\nabla_{\vtheta}\lptheta(\vz,\vx)}%
\global\long\def\lpmargin{\lptheta(\vx)}%
\global\long\def\nlpmargin{\nabla_{\vtheta}\lpmargin}%
\global\long\def\lpjoint{\lptheta(\vz,\vx)}%
\global\long\def\nlpjoint{\ntheta\lpjoint}%
\global\long\def\evalt{\big\rvert_{\vtheta_{t}}}%
\global\long\def\Jtheta{J_{\vtheta}}%
\global\long\def\nJtheta{\ntheta J_{\vtheta}}%
\global\long\def\Jtr{\hat{J}_{\vtheta,\vgamma}}%
\global\long\def\pjoint{p_{\vtheta_{t}}\left(\vzt,\vxt\right)}%
\global\long\def\lpjoint{\lptheta\left(\vzt,\vxt\right)}%
\global\long\def\Dtr{\hat{\Delta}_{\vtheta_{t},\vgamma}}%
\global\long\def\ppostt{\pthetat(\vz|\vx)}%
\global\long\def\natt{\vectorfont{\eta}_{\vtheta}}%
\global\long\def\lnormt{\Phi_{\vtheta}}%
\global\long\def\htr{\hat{\vh}_{\vtheta,\vgamma}}%

\global\long\def\rvZ{\randomvectorfont Z}%
\global\long\def\rvX{\randomvectorfont X}%
\global\long\def\rvZt{\randomvectorfont Z_{t}}%
\global\long\def\rvXt{\randomvectorfont X_{t}}%
\global\long\def\rvZtp{\randomvectorfont Z_{t+1}}%
\global\long\def\rvXtp{\randomvectorfont X_{t+1}}%
\global\long\def\rr{\textnormal{r}}%

\global\long\def\mY{\matrixfont Y}%
\global\long\def\mI{\matrixfont I}%
\global\long\def\mSigma{\matrixfont{\Sigma}}%
\global\long\def\mPhi{\matrixfont{\Phi}}%
\global\long\def\mPsi{\matrixfont{\Psi}}%
\global\long\def\mW{\matrixfont W}%
\global\long\def\mA{\matrixfont A}%
\global\long\def\mC{\matrixfont C}%
\global\long\def\mF{\matrixfont F}%
\global\long\def\mP{\matrixfont P}%
\global\long\def\mU{\matrixfont U}%
\global\long\def\mG{\matrixfont G}%
\global\long\def\mX{\matrixfont X}%
\global\long\def\mM{\matrixfont M}%
\global\long\def\mV{\matrixfont V}%
\global\long\def\mK{\matrixfont K}%
\global\long\def\mB{\matrixfont B}%
\global\long\def\mR{\matrixfont R}%
\global\long\def\me{\matrixfont e}%
\global\long\def\mS{\matrixfont S}%
\global\long\def\mZ{\matrixfont Z}%

\global\long\def\tC{\tensorfont C}%
\global\long\def\tW{\tensorfont W}%
\global\long\def\tG{\tensorfont G}%
\global\long\def\tU{\tensorfont U}%
\global\long\def\tA{\tensorfont A}%

\global\long\def\gD{\graphfont D}%
\global\long\def\D{\gD}%

\global\long\def\Dt{\graphfont D_{t}}%

\global\long\def\Dv{\graphfont D_{v}}%

\global\long\def\gH{\graphfont H}%
\global\long\def\gN{\graphfont N}%
\global\long\def\gU{\graphfont U}%
\global\long\def\gF{\graphfont F}%
\global\long\def\gG{\graphfont G}%
\global\long\def\gQ{\graphfont Q}%
\global\long\def\gR{\graphfont R}%
\global\long\def\gX{\graphfont X}%
\global\long\def\gZ{\graphfont Z}%
\global\long\def\gP{\graphfont P}%
\global\long\def\gS{\graphfont S}%
\global\long\def\gY{\graphfont Y}%
\global\long\def\gC{\graphfont C}%
\global\long\def\gT{\graphfont T}%
\global\long\def\gM{\graphfont M}%
\global\long\def\gI{\graphfont I}%
\global\long\def\gL{\mathcal{L}}%
\global\long\def\gE{\mathcal{E}}%

\global\long\def\sR{\setfont R}%
\global\long\def\sH{\setfont H}%
\global\long\def\sN{\setfont N}%
\global\long\def\sC{\setfont C}%

\global\long\def\E#1#2{\mathbb{E}_{#1}\!\left[#2\right]}%

\global\long\def\argmin{\operatorname*{\arg\min}}%

\global\long\def\argmax{\operatorname*{\arg\max}}%

\global\long\def\arginf{\operatorname*{\arg\inf}}%

\global\long\def\arginf{\operatorname*{\arg\inf}}%

\global\long\def\Dx{\Delta_{\vtheta}(\vx)}%

\global\long\def\Dxt{\Delta_{\vtheta_{t}}(\vx)}%

\global\long\def\KL{\mathrm{KL}}%

\global\long\def\NN{\mathrm{NN}}%

\global\long\def\Dim#1{K_{#1}}%

\global\long\def\iidsim{\stackrel{\text{i.i.d}}{\sim}}%

\global\long\def\tp{^{\intercal}}%

\global\long\def\Tr{\text{Tr}}%

\global\long\def\erva{{\textnormal{a}}}%
 
\global\long\def\ervb{{\textnormal{b}}}%
 
\global\long\def\ervc{{\textnormal{c}}}%
 
\global\long\def\ervd{{\textnormal{d}}}%
 
\global\long\def\erve{{\textnormal{e}}}%
 
\global\long\def\ervf{{\textnormal{f}}}%
 
\global\long\def\ervg{{\textnormal{g}}}%
 
\global\long\def\ervh{{\textnormal{h}}}%
 
\global\long\def\ervi{{\textnormal{i}}}%
 
\global\long\def\ervj{{\textnormal{j}}}%
 
\global\long\def\ervk{{\textnormal{k}}}%
 
\global\long\def\ervl{{\textnormal{l}}}%
 
\global\long\def\ervm{{\textnormal{m}}}%
 
\global\long\def\ervn{{\textnormal{n}}}%
 
\global\long\def\ervo{{\textnormal{o}}}%
 
\global\long\def\ervp{{\textnormal{p}}}%
 
\global\long\def\ervq{{\textnormal{q}}}%
 
\global\long\def\ervr{{\textnormal{r}}}%
 
\global\long\def\ervs{{\textnormal{s}}}%
 
\global\long\def\ervt{{\textnormal{t}}}%
 
\global\long\def\ervu{{\textnormal{u}}}%
 
\global\long\def\ervv{{\textnormal{v}}}%
 
\global\long\def\ervw{{\textnormal{w}}}%
 
\global\long\def\ervx{{\textnormal{x}}}%
 
\global\long\def\ervy{{\textnormal{y}}}%
 
\global\long\def\ervz{{\textnormal{z}}}%

\global\long\def\rvepsilon{{\mathbf{\epsilon}}}%
 
\global\long\def\rvtheta{{\mathbf{\theta}}}%
 
\global\long\def\rva{{\mathbf{a}}}%
 
\global\long\def\rvb{{\mathbf{b}}}%
 
\global\long\def\rvc{{\mathbf{c}}}%
 
\global\long\def\rvd{{\mathbf{d}}}%
 
\global\long\def\rve{{\mathbf{e}}}%
 
\global\long\def\rvf{{\mathbf{f}}}%
 
\global\long\def\rvg{{\mathbf{g}}}%
 
\global\long\def\rvh{{\mathbf{h}}}%
 
\global\long\def\rvu{{\mathbf{i}}}%
 
\global\long\def\rvj{{\mathbf{j}}}%
 
\global\long\def\rvk{{\mathbf{k}}}%
 
\global\long\def\rvl{{\mathbf{l}}}%
 
\global\long\def\rvm{{\mathbf{m}}}%
 
\global\long\def\rvn{{\mathbf{n}}}%
 
\global\long\def\rvo{{\mathbf{o}}}%
 
\global\long\def\rvp{{\mathbf{p}}}%
 
\global\long\def\rvq{{\mathbf{q}}}%
 
\global\long\def\rvr{{\mathbf{r}}}%
 
\global\long\def\rvs{{\mathbf{s}}}%
 
\global\long\def\rvt{{\mathbf{t}}}%
 
\global\long\def\rvu{{\mathbf{u}}}%
 
\global\long\def\rvv{{\mathbf{v}}}%
 
\global\long\def\rvw{{\mathbf{w}}}%
 
\global\long\def\rvx{{\mathbf{x}}}%
 
\global\long\def\rvy{{\mathbf{y}}}%
 
\global\long\def\rvz{{\mathbf{z}}}%

\global\long\def\SDf{\operatorname{SD_{\gF}}}%

\global\long\def\KSDke{\operatorname{\textrm{KSD}_{k,\epsilon}^{2}}}%